%% file: main.tex
\documentclass[10pt,twocolumn,letterpaper]{article}

\usepackage[pagenumbers]{cvpr}
\usepackage{float}
\usepackage{multirow}

\input{preamble}

\definecolor{cvprblue}{rgb}{0.21,0.49,0.74}
\usepackage[pagebackref,breaklinks,colorlinks,citecolor=cvprblue]{hyperref}
\usepackage{color}
\usepackage{gensymb}

\title{GPS-Gaussian: Generalizable Pixel-wise 3D Gaussian Splatting for Real-time Human Novel View Synthesis}

\author{Shunyuan Zheng$^{\dag, 1}$, Boyao Zhou$^2$, Ruizhi Shao$^2$, Boning Liu$^2$, Shengping Zhang$^{*, 1, 3}$,\\ Liqiang Nie$^1$, Yebin Liu$^2$\\
$^1$Harbin Institute of Technology $^2$Tsinghua University $^3$Peng Cheng Laboratory\\
\tt\small \{sawyer0503, s.zhang\}@hit.edu.cn, nieliqiang@gmail.com \\
\tt\small \{bzhou22, liuboning, liuyebin\}@mail.tsinghua.edu.cn, shaorz20@mails.tsinghua.edu.cn}

\begin{document}
\twocolumn[{%
\maketitle
\vspace{-1cm}
\begin{figure}[H]
\hsize=\textwidth %
\centering
\includegraphics[width=\textwidth]{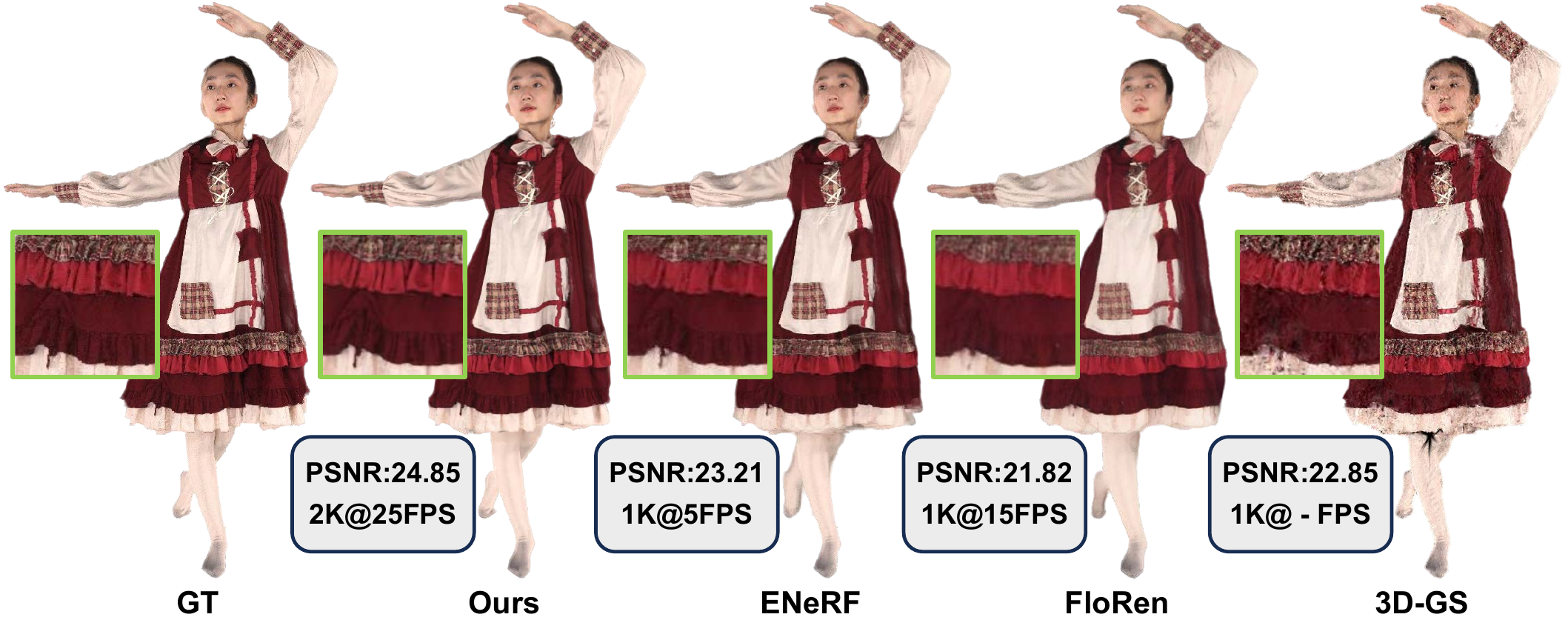}
\vspace{-0.6cm}
\caption{\textbf{High-fidelity and real-time novel view synthesis (NVS).} Our proposed method synthesizes $2K$-resolution novel views of unseen human performers in real-time without any fine-tuning or optimization. The performance outperforms the state-of-the-art feed-forward NVS methods ENeRF~\cite{lin2022enerf}, FloRen~\cite{shao2022floren} and 3D-GS~\cite{kerbl2023_3dgs}, which are representative approaches in Implicit Neural Human Rendering, Image-based Human Rendering and per-subject optimization. We only mark the running efficiency for the feed-forward methods.}
\label{fig:teaser}
\end{figure}
}]

\let\thefootnote\relax\footnotetext{$^\dag$ Work done during an internship at Tsinghua University.}
\let\thefootnote\relax\footnotetext{$^*$ Corresponding author (s.zhang@hit.edu.cn).}

\input{sections/0_abstract}

\input{sections/1_introduction}
\input{sections/2_related}
\input{sections/3_preliminary}

\input{sections/4_1_selection}

\input{sections/4_2_gs_net}

\input{sections/4_3_render}

\input{sections/5_experiments}

\input{sections/6_conclusion}

{
    \small
    \bibliographystyle{ieeenat_fullname}
    \bibliography{main}
}

\input{sections/X_suppl}
\end{document}

%% file: preamble.tex
\usepackage[dvipsnames]{xcolor}


%% file: sections/0_abstract.tex
\begin{abstract}
\vspace{-0.2cm}
We present a new approach, termed GPS-Gaussian, for synthesizing novel views of a character in a real-time manner. 
The proposed method enables 2K-resolution rendering under a sparse-view camera setting.
Unlike the original Gaussian Splatting or neural implicit rendering methods that necessitate per-subject optimizations, we introduce Gaussian parameter maps defined on the source views and regress directly Gaussian Splatting properties for instant novel view synthesis without any fine-tuning or optimization.
To this end, we train our Gaussian parameter regression module on a large amount of human scan data, jointly with a depth estimation module to lift 2D parameter maps to 3D space.
The proposed framework is fully differentiable and experiments on several datasets demonstrate that our method outperforms state-of-the-art methods while achieving an exceeding rendering speed.
The code is available at {\small\url{https://github.com/aipixel/GPS-Gaussian}}.

\end{abstract}

%% file: sections/1_introduction.tex
\vspace{-5mm}
\section{Introduction}
\label{sec:introduction}
Novel view synthesis (NVS) is a critical task that aims to produce photo-realistic images at novel viewpoints from source images captured by multi-view camera systems.
Human NVS, as its subfield, could contribute to 3D/4D immersive scene capture of sports broadcasting, stage performance and holographic communication, which demands real-time efficiency and 3D consistent appearances.
Previous attempts~\cite{chen1993view, oh2001image} synthesize novel views through a weighted blending mechanism~\cite{wilburn2005high}, but they typically rely on dense input views or precise proxy geometry.
Under sparse-view camera settings, it remains a formidable challenge to render high-fidelity images for NVS.

Recently, implicit representations~\cite{saito2019pifu, peng2021neural-body, wang2021neus}, especially Neural Radiance Fields (NeRF)~\cite{mildenhall2020nerf}, have demonstrated remarkable success in numerous NVS tasks.
NeRF utilizes MLPs to represent the radiance field of the scene which jointly predicts the density and color of each sampling point.
To render a specific pixel, the differentiable volume rendering technique is then implemented by aggregating a series of queried points along the ray direction.
The following efforts~\cite{peng2021neural-body, shao2023tensor4d} in human free-view rendering immensely ease the burden of viewpoint quantities while maintaining high qualities.
Despite the progress of accelerating techniques~\cite{muller2022instant-ngp, fridovich2022plenoxels}, NVS methods with implicit representations are time-consuming in general for their dense points querying in scene space.

On the other hand, explicit representations~\cite{peng2021sap, munkberg2022dmtet}, particularly point clouds~\cite{levoy1985use, zwicker2001surface, wiles2020synsin, lassner2021pulsar}, have drawn long-lasting attention due to their high-speed, and even real-time, rendering performance.
Once integrated with neural networks, point-based graphics~\cite{aliev2020npbg, rakhimov2022npbg++} realize a promising explicit representation with comparable realism and extremely superior efficiency in human NVS task~\cite{aliev2020npbg, rakhimov2022npbg++}, compared with NeRF.
More recently, 3D Gaussian Splatting (3D-GS)~\cite{kerbl2023_3dgs} introduces a new representation that the point clouds are formulated as 3D Gaussians with a series of learnable properties including 3D position, color, opacity and anisotropic covariance.
By applying $\alpha$-blending~\cite{kopanas2022neural}, 3D-GS provides not only a more reasonable and accurate mechanism for back-propagating the gradients but also a real-time rendering efficiency for complex scenes.
Despite realizing a real-time inference, Gaussian Splatting relies on a per-subject~\cite{kerbl2023_3dgs} or per-frame~\cite{luiten2023dynamic} parameter optimization for several minutes. 
It is therefore impractical in interactive scenarios as it necessitates the re-optimization of Gaussian parameters once the scene or character changes.

In this paper, we delve into a generalizable 3D Gaussian Splatting method that directly regresses Gaussian parameters in a feed-forward manner instead of per-subject optimization.
Inspired by the success of learning-based human reconstruction, PIFu-like methods~\cite{saito2019pifu, saito2020pifuhd}, we aim to learn the regression of human Gaussian representations from massive 3D human scans with diverse human topologies, clothing styles and pose-dependent deformations.
Deploying these learned human priors, our method enables instantaneous human appearance rendering using a generalizable Gaussian representation.

Specifically, we introduce 2D Gaussian parameter (position, color, scaling, rotation, opacity) maps which are defined on source view image planes, instead of unstructured point clouds.
These Gaussian parameter maps allow us to represent a character with pixel-wise parameters, \textit{i.e.} each foreground pixel corresponding to a specific Gaussian point.
Additionally, it enables the application of efficient 2D convolution networks rather than expensive 3D operators.
To lift 2D parameter maps to 3D Gaussian points, depth maps are estimated for both source views via binocular stereo~\cite{lipson2021raft-stereo} as a learnable unprojection operation.
Such unprojected Gaussian points from both source views constitute the representation of character and novel view images can be rendered with splatting technique~\cite{kerbl2023_3dgs}.

However, the existing cascaded cost volume methods~\cite{shen2021cfnet, lin2022enerf} struggle to tackle the aforementioned depth estimation issue due to the severe self-occlusions in human characters.
Therefore, we propose to learn an iterative stereo-matching~\cite{lipson2021raft-stereo} based depth estimation along with our Gaussian parameter regression, and jointly train the two modules on large-scale data.
Optimal depth estimation contributes to enhanced precision in determining the 3D Gaussian position, while concurrently minimizing rendering loss of Gaussian module rectifies the potential artifacts arising from the depth estimation.
Such a joint training strategy benefits each component and improves the overall stability of the training process.

In practice, we are able to synthesize $2K$-resolution novel views exceeding 25 FPS on a single modern graphics card.
Leveraging the rapid rendering capabilities and broad generalizability inherent in our proposed method, an unseen character can be instantly rendered without necessitating any fine-tuning or optimization, as illustrated in Fig.~\ref{fig:teaser}.
In summary, our contributions can be summarized as follows:
\noindent
\begin{itemize}
    \item We introduce a generalizable 3D Gaussian Splatting methodology that employs pixel-wise Gaussian parameter maps defined on 2D source image planes to formulate 3D Gaussians in a feed-forward manner.

    \item We propose a fully differentiable framework composed of an iterative depth estimation module and a Gaussian parameter regression module. The intermediate predicted depth map bridges the two components and allows them to benefit from joint training.

    \item We develop a real-time NVS system that achieves $2K$-resolution rendering by directly regressing Gaussian parameter maps. 

\end{itemize}

%% file: sections/2_related.tex
\section{Related Work}
\paragraph{Neural Implicit Human Representation.}
Neural implicit function has recently aroused a surge of interest to represent complicated scenes, in form of occupancy fields~\cite{mescheder2019occupancy, saito2019pifu, saito2020pifuhd, hong2021stereopifu}, neural radiance fields~\cite{mildenhall2020nerf, peng2021neural-body, zhao2022humannerf, weng2022humannerf, guo2023vid2avatar} and neural signed distance functions~\cite{park2019deepsdf, wang2021neus, wang2023neus2, shao2023tensor4d, zhou2023human}.
Implicit representation shows the advantage in memory efficiency and topological flexibility for human reconstruction task~\cite{hong2021stereopifu, zheng2021pamir, xiu2022icon}, especially in a pixel-aligned feature query manner~\cite{saito2019pifu, saito2020pifuhd}.
However, each queried point is processed through the full network, which dramatically increases computational complexity.  
More recently, numerous methods have extended Neural Radiance Fields (NeRF)~\cite{mildenhall2020nerf} to static human modeling~\cite{shao2022doublefield, chen2023gm-nerf} and dynamic human modeling from sparse multi-view cameras~\cite{peng2021neural-body, zhao2022humannerf, shao2023tensor4d} or a monocular camera~\cite{weng2022humannerf, jiang2022neuman, guo2023vid2avatar}.
However, these methods typically require a per-subject optimization process and it is non-trivial to generalize these methods to unseen subjects. Previous attempts, \textit{e.g.}, PixelNeRF~\cite{yu2021pixelnerf}, IBRNet~\cite{wang2021ibrnet}, MVSNeRF~\cite{chen2021mvsnerf} and ENeRF~\cite{lin2022enerf} resort to image-based features as potent prior cues for feed-forward scene modeling.
The large variation in pose and clothing makes generalizable NeRF for human rendering a more challenging task, thus recent work simplifies the problem by leveraging human priors.
For example, NHP~\cite{kwon2021nhp}, GM-NeRF~\cite{chen2023gm-nerf} and TransHuman~\cite{pan2023transhuman} employ parametric human body model (SMPL~\cite{loper2015smpl}), KeypointNeRF~\cite{mihajlovic2022keypointnerf} uses 3D skeleton keypoints to encode spatial information. These additional processes increase computational cost and an inaccurate prior estimation would mislead the final result.
On the other hand, despite the great progress in accelerating the scene-specific NeRF~\cite{yu2021plenoctrees, fridovich2022plenoxels, muller2022instant-ngp, li2023nerfacc}, efficient generalizable NeRF for interactive scenarios remains to be further elucidated.

\paragraph{Deep Image-based Rendering.}
Image-based rendering, or IBR in short, synthesizes novel views from a set of multi-view images with a weighted blending mechanism, which is typically computed from a geometry proxy.
\cite{riegler2020free, riegler2021stable} deploy multi-view stereo from dense input views to produce mesh surfaces as a proxy for image warping.
DNR~\cite{thies2019dnr} directly produces learnable features on the surface of mesh proxies for neural rendering.
Obtaining these proxies is not straightforward since high-quality multi-view stereo and surface reconstruction requires dense input views.
MonoFVV~\cite{guo2017real}, LookinGood~\cite{martin2018lookingood} and Function4D~\cite{yu2021function4d} implement RGBD fusion to attain real-time human rendering.
Point clouds from SfM~\cite{meshry2019neural, pittaluga2019revealing} or depth sensors~\cite{nguyen2022hsv-net} can also be engaged as geometry proxies.
These methods highly depend on the performance of 3D reconstruction algorithms or the quality of depth sensors. 
FWD~\cite{cao2022fwd} designs a network to refine depth estimations, then explicitly warps pixels from source views to novel views with the refined depth maps.
FloRen~\cite{shao2022floren} utilizes a coarse human mesh reconstructed by PIFu~\cite{saito2019pifu} to render initialized depth maps for novel views.
Arguably most related to ours is FloRen~\cite{shao2022floren}, as it also realizes $360^\circ$ free view human performance rendering in real-time.
However, the appearance flow in FloRen merely works in 2D domains, where the rich geometry cues and multi-view geometric constraints only serve as 2D supervisions.
The difference is that our approach lifts 2D priors into 3D space and utilizes the point representation to synthesize novel views in a fully differentiable manner.

\paragraph{Point-based Graphics.}
Point-based representation has shown great efficiency and simplicity for various 3D human tasks~\cite{liu2009point, zhou2020reconstructing, ma2021power, lin2022eccv, zheng2023pointavatar, zhang2023closet}.
Previous attempts integrate point cloud representation with 2D neural rendering~\cite{aliev2020npbg, rakhimov2022npbg++} or NeRF-like volume rendering~\cite{xu2022point-nerf, su2023npc}.
Still, such a hybrid architecture does not exploit rendering capability of point cloud and takes a long time to optimize on different scenes.
Then differentiable point-based~\cite{wiles2020synsin} and sphere-based~\cite{lassner2021pulsar} rendering have been developed, which demonstrates promising rendering qualities, especially attaching them to a conventional network pipeline~\cite{cao2022fwd, nguyen2022hsv-net}. 
In addition, isotropic points can be substituted by a more reasonable Gaussian point modeling~\cite{kerbl2023_3dgs, luiten2023dynamic} to realize a rapid differentiable rendering framework with a splatting technique. 
This advanced representation has showcased prominent performance in concurrent 3D human work~\cite{hu2024gaussianavatar, shao2024control4d, xu2024gaussian, li2024animatable, liu2024humangaussian}.
However, a per-scene or per-subject optimization strategy limits its real-world application.
In this paper, we go further to generalize 3D Gaussians across diverse subjects while maintaining its fast and high-quality rendering properties.

%% file: sections/3_preliminary.tex
\section{Preliminary}
\label{sec:preliminary} 
Since the proposed GPS-Gaussian harnesses the power of 3D-GS~\cite{kerbl2023_3dgs}, we give a brief introduction in this section.

3D-GS models a static 3D scene explicitly with point primitives, each of which is parameterized as a scaled Gaussian with 3D covariance matrix $\mathbf{\Sigma}$ and mean $\mathbf{\mu}$
\begin{equation}
\label{formula:gaussian formula}
    G(\mathcal{X})~= e^{-\frac{1}{2}(\mathcal{X}-\mu)^{T}{\mathbf{\Sigma}}^{-1}(\mathcal{X}-\mu)}
\end{equation}
In order to be effectively optimized by gradient descent, the covariance matrix $\mathbf{\Sigma}$ can be decomposed into a scaling matrix $\mathbf{S}$ and a rotation matrix $\mathbf{R}$ as
\begin{equation}
\label{formula:covariance decomposition}
    \mathbf{\Sigma} = \mathbf{RSS}^T\mathbf{R}^T
\end{equation}
Following~\cite{zwicker2002ewa}, the projection of Gaussians from 3D space to a 2D image plane is implemented by a view transformation $\mathbf{W}$ and the Jacobian of the affine approximation of the projective transformation $\mathbf{J}$. The covariance matrix ${\mathbf{\Sigma}}^{\prime}$ in 2D space can be computed as
\begin{equation}
    \mathbf{\Sigma^{\prime}} = \mathbf{JW\Sigma W}^T\mathbf{J}^T
\end{equation}
followed by a point-based alpha-blend rendering which bears similarities to that used in NeRF~\cite{mildenhall2020nerf}, formulated as
\begin{equation}
\label{formula: splatting&volume rendering}
    \mathbf{C}_{color} = \sum_{i\in N}\mathbf{c}_i \alpha_i \prod_{j=1}^{i-1} (1-\alpha_i)
\end{equation}
where $\mathbf{c}_i$ is the color of each point, and density $\alpha_i$ is reasoned by the multiplication of a 2D Gaussian with covariance ${\mathbf{\Sigma}}^{\prime}$ and a learned per-point opacity~\cite{yifan2019differentiable}. The color is defined by spherical harmonics (SH) coefficients in \cite{kerbl2023_3dgs}.

To summarize, the original 3D Gaussians methodology characterizes each Gaussian point by the following attributes: (1) a 3D position of each point $\mathcal{X} \in \mathbb{R}^3$, (2) a color defined by SH $\mathbf{c} \in \mathbb{R}^k$ (where $k$ is the freedom of SH basis), (3) a rotation parameterized by a quaternion $\mathbf{r} \in \mathbb{R}^4$, (4) a scaling factor $\mathbf{s} \in \mathbb{R}_+^3$, and (5) an opacity $\alpha \in [0, 1]$.

%% file: sections/4_1_selection.tex
\begin{figure*}
  \centering
  \includegraphics[width=\textwidth]{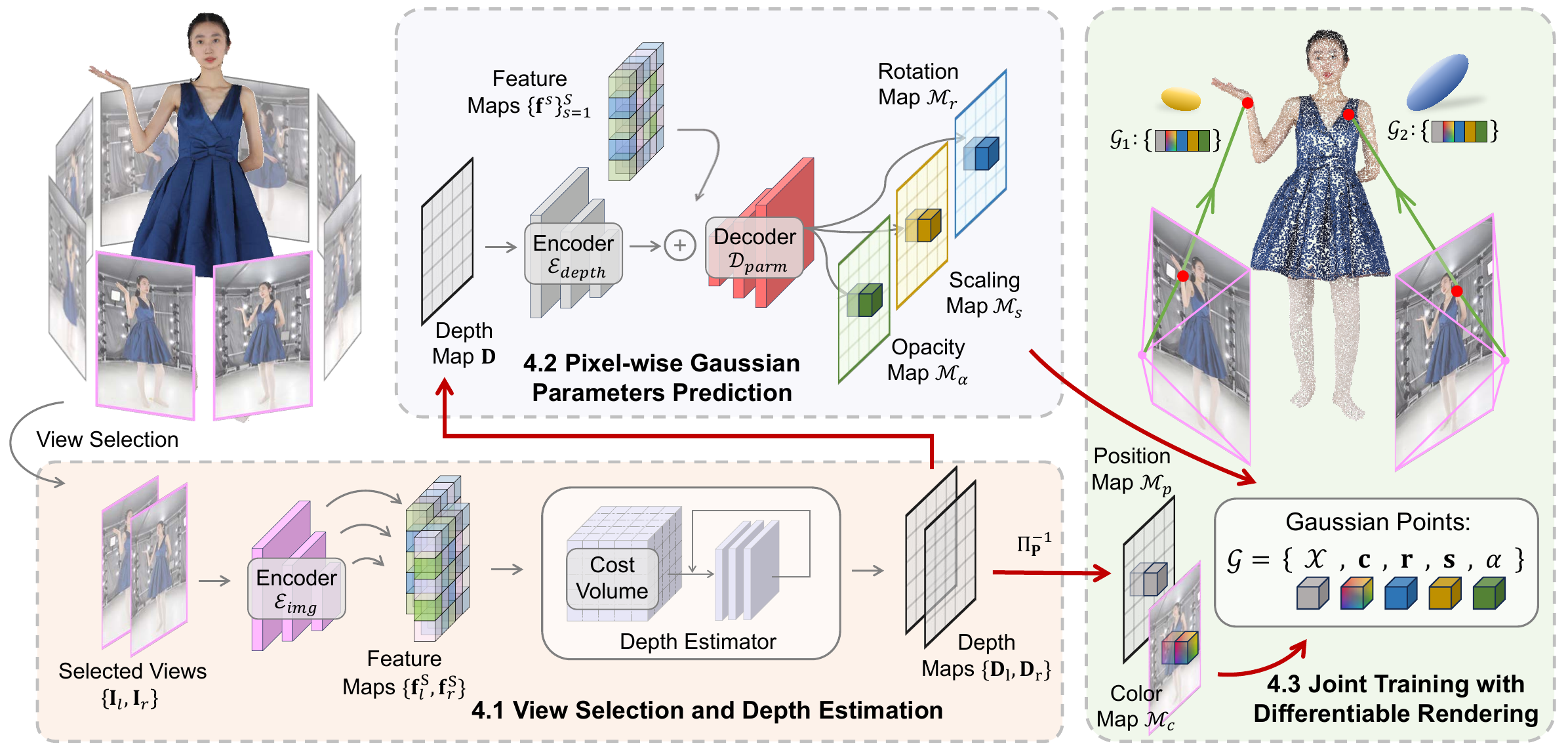}
  \vspace{-5mm}
  \caption{\textbf{Overview of GPS-Gaussian.} Given RGB images of a human-centered scene with sparse camera views and a target novel viewpoint, we select the adjacent two views on which to formulate our Gaussian representation. We extract the image features followed by conducting an iterative depth estimation. For each source view, the depth map and the RGB image serve as a 3D position map and a color map, respectively, to formulate the Gaussian representation while the other parameters of 3D Gaussians are predicted in a pixel-wise manner. The Gaussian parameter maps defined on 2D image planes of both views are further unprojected to 3D space and aggregated for novel view rendering. The fully differentiable framework enables a joint training mechanism for all networks.}
  \vspace{-2mm}
  \label{fig:pipeline}
\end{figure*}

\section{Method}
\label{sec:method}

The overview of our method is illustrated in Fig.~\ref{fig:pipeline}.
Given the RGB images of a human-centered scene with sparse camera views, our method aims to generate high-quality free-viewpoint renderings of the performer in real-time.
Once given a target novel viewpoint, we select the two neighboring views and extract the image features using a shared image encoder.
Following this, a binocular depth estimator takes the extracted features as input to predict the depth maps for both source views (Sec.~\ref{sec:selection&depth}).
The depth values and the RGB values in foreground regions of the source view determine the 3D position and color of each Gaussian point, respectively, while the other parameters of 3D Gaussians are predicted in a pixel-wise manner (Sec.~\ref{sec:gaussian_network}).
Combined with the depth map and source RGB image, these parameter maps formulate the Gaussian representation in 2D image planes and are further unprojected to 3D space.
The unprojected Gaussians from both views are aggregated and rendered to the target viewpoint in a differentiable way, which allows for end-to-end training (Sec.~\ref{sec:render}).

\subsection{View Selection and Depth Estimation}
\label{sec:selection&depth}
\noindent\textbf{View Selection.}
Unlike the original 3D Gaussians that optimize the characteristics of each Gaussian point on all source views, we synthesize the desired novel view with two adjacent source views.
Given $N$ input images $\{\mathbf{I}_n\}_{n=1}^N$, with their camera position $\{C_n\}_{n=1}^N$, source views can be represented by $\mathbf{V_n}=C_n-O$, where $O$ is the center of the scene.
Similarly, the target novel view rendering can be defined as $I_{tar}$ with camera position $C_{tar}$ and view $\mathbf{V_{tar}}=C_{tar}-O$.
By conducting a dot product of all input views vectors and the novel view vector, the nearest two views $(v_{l}, v_{r})$ can be selected as the `working set' of binocular stereo, where $l$ and $r$ stand for `left' and `right' view, respectively.

The rectified source images $\mathbf{I}_{l}, \mathbf{I}_{r}\in[0,1]^{H\times W\times 3}$ are fed to a shared image encoder $\mathcal{E}_{img}$ with several residual blocks and downsampling layers to extract dense feature maps $\mathbf{f}^s \in \mathbb{R}^{H/2^s\times W/2^s\times D_s}$ where $D_s$ is the dimension at the $s$-th feature scale
\begin{equation}
\label{formula:image encoder}
    \langle\{\mathbf{f}_l^s\}_{s=1}^S, \{\mathbf{f}_r^s\}_{s=1}^S\rangle = \mathcal{E}_{img}(\mathbf{I}_{l}, \mathbf{I}_{r})
\end{equation}
where we set $S=3$ in our experiments.

\noindent\textbf{Depth Estimation.}
The depth map is the key component of our framework bridging the 2D image planes and 3D Gaussian representation.
Note that, depth estimation in binocular stereo is equivalent to disparity estimation.
For each pixel $(u, v)$ in one view, disparity estimation $\phi_{disp}$ aims to find its corresponding coordinate $(u+\phi_{disp}(u), v)$ in another view, considering the displacement of each pixel is constrained to a horizontal line in rectified stereo.
Since the predicted disparity maps can be easily converted to depth maps given camera parameters, we do not distinguish them in the following sections.
In theory, any alternative depth estimation methods can be adapted to our framework.
We implement this module in an iterative manner inspired by \cite{lipson2021raft-stereo} mainly because it avoids using prohibitively slow 3D convolutions to filter the cost volume.

Given the feature maps $\mathbf{f}_l^S, \mathbf{f}_r^S \in \mathbb{R}^{H/2^S\times W/2^S\times D_S}$, we compute a 3D correlation volume $\mathbf{C} \in \mathbb{R}^{H/2^S\times W/2^S\times W/2^S}$ using a matrix multiplication
\begin{equation}
\label{formula:cost volume}
    \mathbf{C}(\mathbf{f}_l^S, \mathbf{f}_r^S), \quad C_{ijk} = \sum_h (\mathbf{f}_l^S)_{ijh} \cdot (\mathbf{f}_r^S)_{ikh}
    \vspace{-2mm}
\end{equation}
Then, an iterative update mechanism predicts a sequence of depth estimations $\{\mathbf{d}_l^t\}_{t=1}^T$ and $\{\mathbf{d}_r^t\}_{t=1}^T$ by looking up in volume $\mathbf{C}$, where $T$ is the update iterations.
For more details about the update operators, please refer to \cite{teed2020raft}.
The outputs of final iterations ($\mathbf{d}_l^{T}$, $\mathbf{d}_r^{T}$) are upsampled to full image resolution via a convex upsampling.
The depth estimation module $\Phi_{depth}$ can be formulated as
\begin{equation}
\label{formula:disp estimation}
    \langle\mathbf{D}_l, \mathbf{D}_r\rangle = \Phi_{depth}(\mathbf{f}_l^S, \mathbf{f}_r^S, K_l, K_r)
\end{equation}
where $K_l$ and $K_r$ are the camera parameters, $\mathbf{D}_l, \mathbf{D}_r \in \mathbb{R}^{H \times W \times 1}$ are the depth estimations.
The classic binocular stereo methods estimate the depth for `reference views' only, while we pursue depth maps of both inputs to formulate the Gaussian representation, which makes our implementation highly symmetrical.
By leveraging this nature, we realize a compact and highly parallelized module that results in a decent efficiency increase.
Detailed designs of this module can be seen in our supplementary material.

%% file: sections/4_2_gs_net.tex
\subsection{Pixel-wise Gaussian Parameters Prediction}
\label{sec:gaussian_network}
Each Gaussian point in 3D space is characterized by attributes $\mathcal{G}=\{\mathcal{X}, \mathbf{c}, \mathbf{r}, \mathbf{s}, \alpha\}$, which represent 3D position, color, rotation, scaling and opacity, respectively.
In this section, we introduce a pixel-wise manner to formulate 3D Gaussians in 2D image planes.
Specifically, the proposed Gaussian maps $\mathbf{G}$ are defined as
\begin{equation}
\label{formula:gs_def}
    \mathbf{G}(x) = \{\mathcal{M}_p(x), \mathcal{M}_c(x), \mathcal{M}_r(x), \mathcal{M}_s(x), \mathcal{M}_\alpha(x)\}
\end{equation}
where $x$ is the coordinate of a foreground pixel in an image plane, $\mathcal{M}_p, \mathcal{M}_c, \mathcal{M}_r, \mathcal{M}_s, \mathcal{M}_\alpha$ represents Gaussian parameter maps of position, color, rotation, scaling and opacity, respectively.
Given the predicted depth map $\mathbf{D}$, a pixel located at $x$ can be immediately unprojected from image planes to 3D space using projection matrix $\mathbf{P} \in \mathbb{R}^{3 \times 4} $ structure with camera parameters $K$
\begin{equation}
\label{formula:gs_pisition}
    \mathcal{M}_p(x) = \Pi^{-1}_\mathbf{P}(x, \mathbf{D}(x))
\end{equation}
Thus the learnable unprojection in Eq.~\ref{formula:gs_pisition} bridges 2D feature space and 3D Gaussian representation.
Considering our human-centered scenario is predominantly characterized by diffuse reflection, instead of predicting the SH coefficients, we directly use the source RGB image as the color map
\vspace{-1mm}
\begin{equation}
\label{formula:gs_color}
    \mathcal{M}_c(x) = \mathbf{I}(x)
    \vspace{-1mm}
\end{equation}
We argue that the remaining three Gaussian parameters are generally related to (1) pixel level local features, (2) the global context of human bodies, and (3) detailed spatial structures.
Image features $\{\mathbf{f}^s\}_{s=1}^S$ from encoder $\mathcal{E}_{img}$ have already derived strong cues of (1) and (2).
Hence, we construct an additional encoder $\mathcal{E}_{depth}$, which takes the depth map $\textbf{D}$ as input, to complement the geometric awareness for each pixel.
The image features and the spatial features are fused by a U-Net like decoder $\mathcal{D}_{parm}$ to regress pixel-wise Gaussian features in full image resolution
\begin{equation}
\label{formula:decoder}
    \mathbf{\Gamma} = \mathcal{D}_{parm} (\mathcal{E}_{img}(\mathbf{I}) \oplus\mathcal{E}_{depth}(\mathbf{D}))
\end{equation}
where $\mathbf{\Gamma} \in \mathbb{R}^{H \times W \times D_G}$ is Gaussian features, $\oplus$ stands for concatenations at all feature levels.
The prediction heads, each composed of 2 convolution layers, are adapted to Gaussian features for specific Gaussian parameter map regression.
Before being used to formulate Gaussian representations, the rotation map should be normalized since it represents a quaternion
\begin{equation}
\label{formula:gs_rotation}
    \mathcal{M}_r(x) = Norm({h}_r(\mathbf{\Gamma}(x)))
    \vspace{-1mm}
\end{equation}
where ${h}_r$ is the rotation head.
The scaling map and the opacity map need activations to satisfy their range
\begin{align}
\begin{split}
\label{formula:gs_scaling&opacity}
    \mathcal{M}_s(x) = Softplus({h}_s(\mathbf{\Gamma}(x))) \\
    \mathcal{M}_\alpha(x) = Sigmoid({h}_\alpha(\mathbf{\Gamma}(x)))
\end{split}
\end{align}
where ${h}_s$ and ${h}_\alpha$ represent the scaling head and opacity head, respectively.
The detailed network architecture in this section is provided in our supplementary material.

%% file: sections/4_3_render.tex
\input{tables/quantitative}

\tablecompare

\subsection{Joint Training with Differentiable Rendering}
\label{sec:render}

The pixel-wise Gaussian parameter maps defined on both source views are then lifted to 3D space and aggregated to render photo-realistic novel view images using the Gaussian Splatting technique in Sec.~\ref{sec:preliminary}.

\noindent\textbf{Joint Training Mechanism.} The fully differentiable rendering framework simultaneously enables joint training from two perspectives:
(1) The depth estimations of both source views.
(2) The depth estimation module and the Gaussian parameter prediction module.
As for the former, the independent training of depth estimators on two source views makes the 3D representation inconsistent due to the mismatch of the source views.
As for the latter, the classic stereo-matching based depth estimation is fundamentally a 2D task that aims at densely finding the correspondence between pixels from two images. The differentiable rendering integrates auxiliary 3D awareness.
On the other hand, optimal depth estimation contributes to enhanced precision in determining the 3D Gaussian parameters. 

\noindent\textbf{Loss Functions.} We use L1 loss and SSIM loss~\cite{wang2004ssim}, denoted as $\mathcal{L}_{mae}$ and $\mathcal{L}_{ssim}$ respectively, to measure the difference between the rendered and ground truth image 
\begin{equation}
    \mathcal{L}_{render} = \beta{L}_{mae} + \gamma\mathcal{L}_{ssim}
\end{equation}
where we set $\beta=0.8$ and $\gamma=0.2$ in our experiments.
Similar to \cite{lipson2021raft-stereo}, we supervise on the L1 distance between the predicted and ground truth depth over the full sequence of predictions $\{\mathbf{d}^t\}_{t=1}^T$ with exponentially increasing weights.
Given ground truth depth $\mathbf{d}_{gt}$, the loss is defined as
\begin{equation}
    \mathcal{L}_{disp} = \sum_{t=1}^{T} \mu^{T-t} \|\mathbf{d}_{gt} - \mathbf{d}^t\|_1
\end{equation}
where we set $\mu=0.9$ in our experiments.
Our final loss function is $\mathcal{L} = \mathcal{L}_{render} + \mathcal{L}_{disp}$.

%% file: tables/quantitative.tex
\newcommand{\tablecompare}{
\begin{table*}
\small
\caption{\textbf{Quantitative comparison on THuman2.0~\cite{yu2021function4d}, Twindom~\cite{twindom} and our collected real-world data.} All methods are evaluated on an RTX 3090 GPU to report the speed of synthesizing one novel view with two $1024 \times 1024$ source images. Our method and FloRen~\cite{shao2022floren} use TensorRT for fast inference. $\dagger$ 3D-GS~\cite{kerbl2023_3dgs} requires per-subject optimization, while the other methods perform feed-forward inferences.}
\vspace{-1mm}
\centering
\begin{tabular}{l|ccc|ccc|ccc|c}
\toprule
\multicolumn{1}{c}{\multirow{1}[6]{*}{Method}} & \multicolumn{3}{c}{THuman2.0~\cite{yu2021function4d}} & \multicolumn{3}{c}{Twindom~\cite{twindom}} & \multicolumn{3}{c}{Real-world Data} & \multirow{1}[6]{*}{FPS}\\

\cmidrule{2-10} \multicolumn{1}{c}{} & PSNR$\uparrow$ & SSIM$\uparrow$ & \multicolumn{1}{c}{LPIPS$\downarrow$} & PSNR$\uparrow$ & SSIM$\uparrow$  & \multicolumn{1}{c}{LPIPS$\downarrow$} & PSNR$\uparrow$ & SSIM$\uparrow$ & \multicolumn{1}{c}{LPIPS$\downarrow$} \\ \midrule

3D-GS~\cite{kerbl2023_3dgs}$\dagger$     & 24.18 & 0.821 & 0.144 & 22.77 & 0.785 & 0.153 & 22.97 & 0.839 & 0.125 & /\\
FloRen~\cite{shao2022floren}   & 23.26 & 0.812 & 0.184 & 22.96 & 0.838 & 0.165 & 22.80 & 0.872 & 0.136 & 15\\
IBRNet~\cite{wang2021ibrnet}  & 23.38 & 0.836 & 0.212 & 22.92 & 0.803 & 0.238 & 22.63 & 0.852 & 0.177 & 0.25\\
ENeRF~\cite{lin2022enerf}     & 24.10 & 0.869 & 0.126 & 23.64 & 0.847 & 0.134 & 23.26 & 0.893 & 0.118 & 5\\
Ours     & \textbf{25.57} & \textbf{0.898} & \textbf{0.112} & \textbf{24.79} & \textbf{0.880} & \textbf{0.125} & \textbf{24.64} & \textbf{0.917} & \textbf{0.088} & \textbf{25}\\
\bottomrule
\end{tabular}
\vspace{-2mm}
\label{tab:num_compare}
\end{table*}
}

%% file: sections/5_experiments.tex
\input{tables/camera_sparsity}

\input{tables/ablation}

\begin{figure*}
  \centering
  \includegraphics[width=\textwidth]{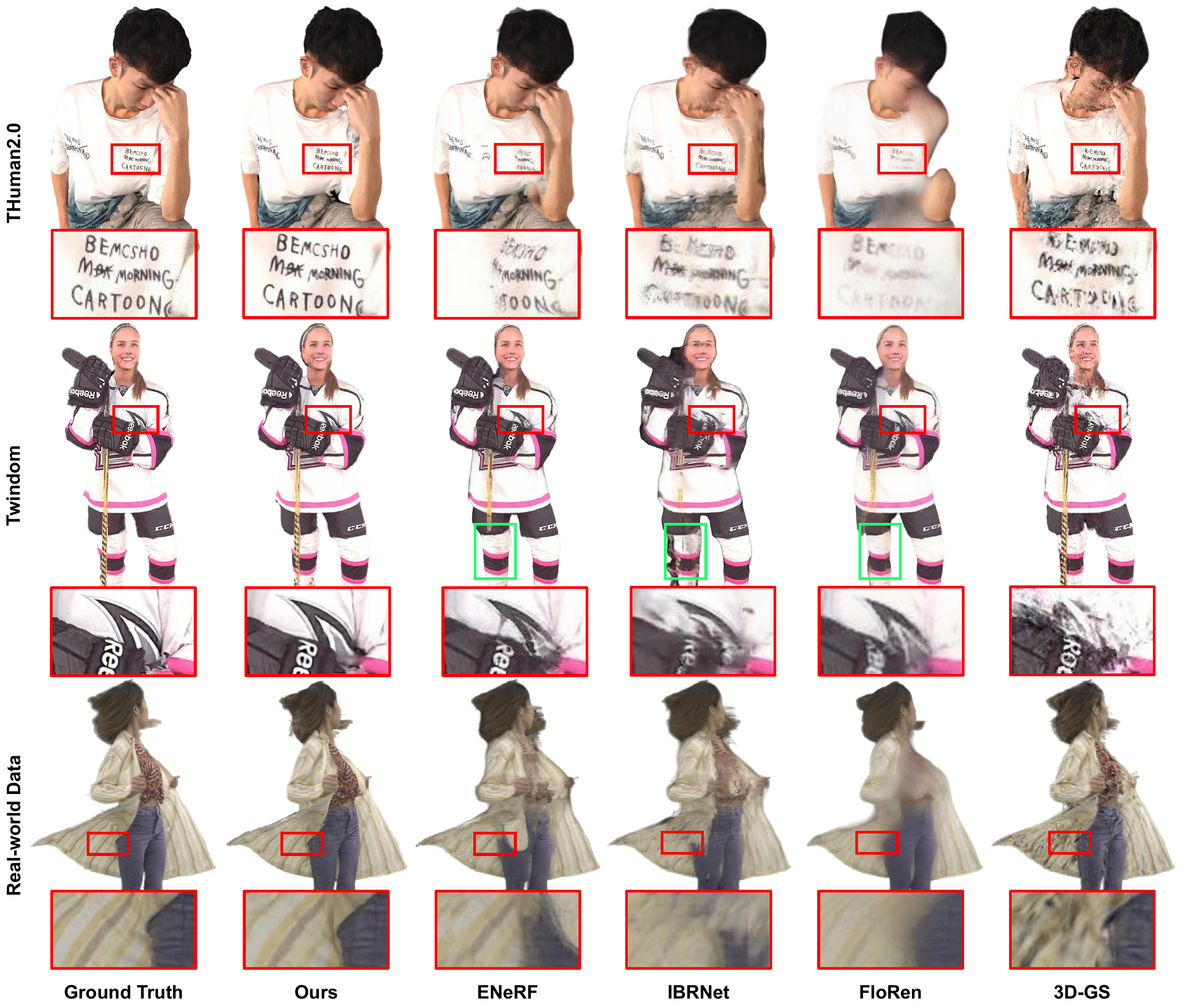}
  \caption{\textbf{Qualitative comparison on THuman2.0~\cite{yu2021function4d}, Twindom~\cite{twindom} and our collected real-world data.} Our method produces more detailed human appearances and can recover more reasonable geometry.}
  \label{fig:comparison}
  \vspace{-2mm}
\end{figure*}

\section{Experiments}
\label{sec:experiments}

\subsection{Implementation Details} 
Our GPS-Gaussian is trained on a single RTX3090 graphics card using AdamW~\cite{adamw} optimizer with an initial learning rate of $2e^{-4}$.
Since the unstable depth estimation in the very first training steps can have a strong impact on Gaussian parameter regression, we pre-train the depth estimation module for $40k$ iterations.
Then we jointly train two modules for $100k$ iterations with a batch size of $2$ and the overall training process takes around 15 hours. 

\subsection{Datasets and Metrics} 
To learn human priors from a large amount of data, we collect 1700 and 526 human scans from Twindom~\cite{twindom} and THuman2.0~\cite{yu2021function4d}, respectively.
We randomly select 200 and 100 scans as validation data from Twindom and THuman2.0, respectively.
As shown in Fig.~\ref{fig:pipeline}, we uniformly position 8 cameras in a cycle, thus the angle between two neighboring cameras is about $45^{\circ}$. 
We render synthetic human scans to these camera positions as source view images while randomly choosing 3 viewpoints to render novel view images, which are positioned on the intersection arc between each two adjacent input views.
To test the robustness in real-world scenarios, we capture real data of 4 characters in the same 8-camera setup and prepare 8 additional camera views for evaluation. 
Similar to ENeRF~\cite{lin2022enerf}, we evaluate PSNR, SSIM~\cite{wang2004ssim} and LPIPS~\cite{zhang2018lpips} as metrics for the rendering results in foreground regions determined by the bounding box of humans. 

\subsection{Comparisons with State-of-the-art Methods}
\noindent\textbf{Baselines.}
Considering that our goal is instant novel view synthesis, we compare our GPS-Gaussian against three generalizable methods including implicit method ENeRF~\cite{lin2022enerf}, image-based rendering method FloRen~\cite{shao2022floren} and hybrid method IBRNet~\cite{wang2021ibrnet}. 
All baseline methods are trained from scratch on the same dataset as ours and take two source views as input for synthesizing the targeted novel view.
Note that, our method and FloRen use ground truth depths for supervision.
We further prepare the comparison with the original 3D-GS~\cite{kerbl2023_3dgs} which is optimized on all 8 input views using the default strategies in the released code.

\noindent\textbf{Comparison Results.}
The comparisons on both synthetic and real-world data are listed in Table~\ref{tab:num_compare}. Our GPS-Gaussian outperforms all methods on all metrics and achieves a much faster rendering speed. 
Qualitative rendering results in Fig.~\ref{fig:comparison} show that our method can synthesize fine-grained novel view images with more detailed appearances.
Once occlusion happens, some target regions under the novel view are invisible in one or both of the source views.
The resulting depth ambiguity between input views causes ENeRF and IBRNet to render unreasonable results since these methods are confused when conducting the feature aggregation.
The unreliable geometric proxy in these cases also makes FloRen produce blurred outputs even if it employs the depth and flow refining networks.
In our method, the human priors learned from massive human images help to alleviate the adverse effects caused by occlusion.
In addition, 3D-GS takes several minutes for optimization and produces noisy rendering results of novel views in such a sparse camera setup.
Also, most of the compared methods have difficulty in handling thin structures such as hockey sticks and robes in Fig.~\ref{fig:comparison}.
We further prepare the sensitivity analysis of camera view sparsity in Table~\ref{tab:camera}.
For 6-camera results, we use the same models trained under 8-camera setup without any fine-tuning.
Among baselines, our method degrades reasonably and holds robustness when decreasing cameras.
We ignore 3D-GS here because it takes several minutes for per-subject optimization and produces noisy rendering results, as shown in Fig.~\ref{fig:comparison}, even in 8-camera setup.

\tabledegree

\subsection{Ablation Studies}
\label{sec:ablation}

We evaluate the effectiveness of our designs in more detail through ablation experiments.
Other than rendering metrics, we follow \cite{lipson2021raft-stereo} to evaluate depth (identical to disparity) estimation with the end-point-error (EPE) and the ratio of pixel error in 1 pix level.
All ablations are trained and tested on the aforementioned synthetic data.

\tableablation

\begin{figure}
  \centering
  \vspace{-1mm}
  \includegraphics[width=\linewidth]{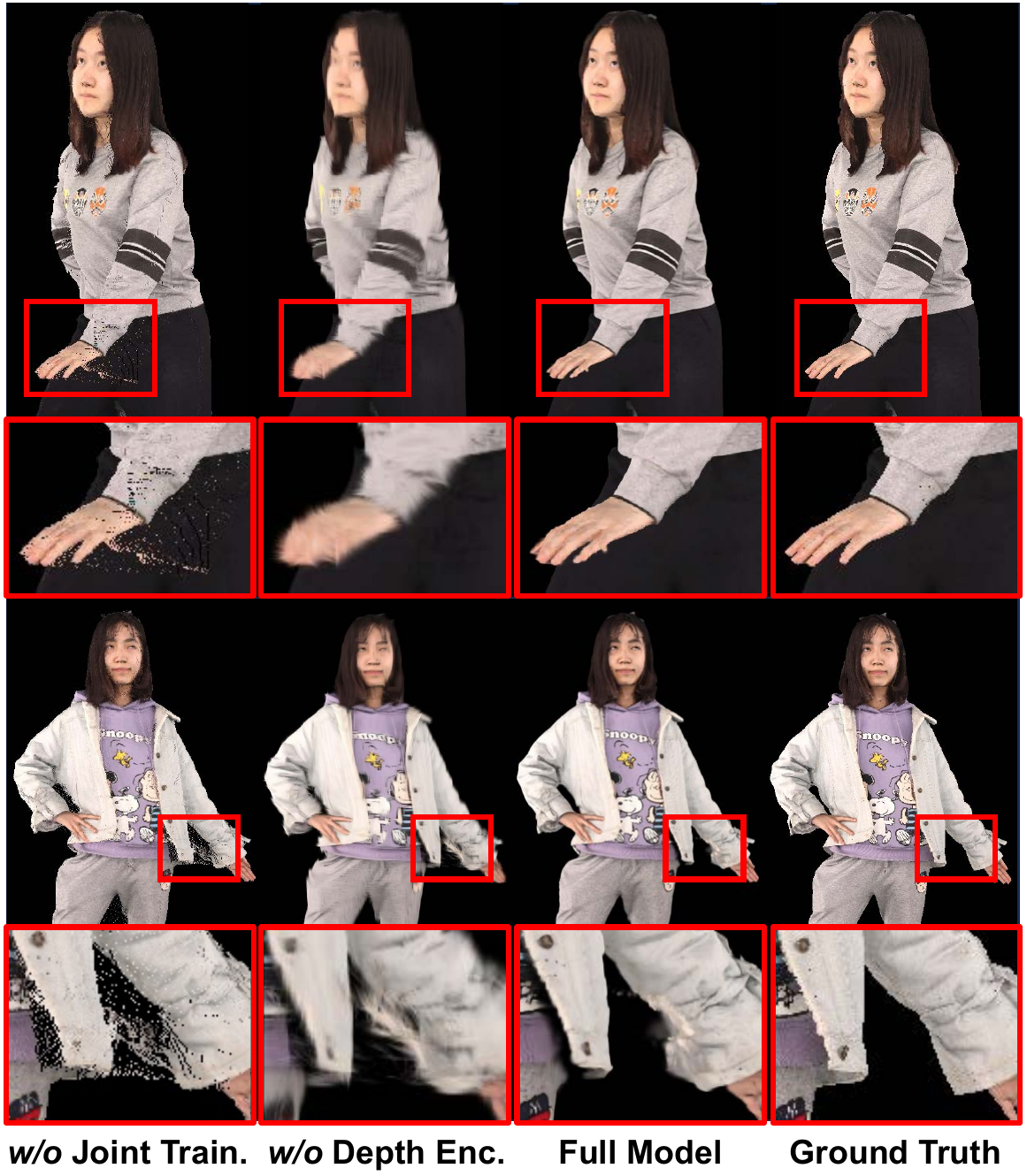}
  \vspace{-5mm}
  \caption{\textbf{Qualitative ablation study on synthetic data.} We show the effectiveness of the joint training and the depth encoder in the full pipeline. 
  The proposed designs make the rendering results more visually appealing with fewer artifacts and less blurry.}
  \label{fig:ablation}
  \vspace{-4mm}
\end{figure}

\noindent\textbf{Effects of Joint Training Mechanism.}
We design a model without the differentiable Gaussian rendering by substituting it with point cloud rendering at a fixed radius.
Thus the model degenerates into a depth estimation network and an undifferentiable depth warping based rendering. 
The rendering quality is merely based on the accuracy of depth estimation while the rendering loss could not conversely promote the depth estimator.
We train the ablated model for the same iterations as the full model for fair comparison.
The rendering results in Fig.~\ref{fig:ablation} witness obvious noise due to the depth ambiguity in the margin area of the source views where the depth value changes drastically.
The rendering noise causes a degradation in PSNR and SSIM as manifested in Table~\ref{tab:ablation}, while it cannot be reflected in the perception metric LPIPS.
The joint regression with Gaussian parameters precisely recognizes these outliers and compensates for these artifacts by predicting an extremely low opacity for the Gaussian points centered at these positions.
Please refer to the supplementary material for the visualization of opacity maps.
Meanwhile, the independent training of the depth estimation module interrupts the interaction of two source views, resulting in an inconsistent geometry.
As illustrated in Table~\ref{tab:ablation}, joint training makes a more robust depth estimator with a $5\%$ improvement in EPE.

\noindent\textbf{Effects of Depth Encoder.}
We claim that merely using image features is insufficient for predicting Gaussian parameters.
Herein, we ablate the depth encoder from our full model, thus the Gaussian parameter decoder only takes as input the image features to predict $\mathcal{M}_r, \mathcal{M}_s, \mathcal{M}_\alpha$ simultaneously.
As shown in Fig.~\ref{fig:ablation}, the ablated model fails to recover the details of human appearance, leading to blurred rendering results.
The scale of Gaussian points is impacted by comprehensive factors including depth, texture and surface roughness.
The absence of spatial awareness degrades the regression of scaling map $\mathcal{M}_s$, which deteriorates the visual perception reflected on LPIPS, even with a comparable depth estimation accuracy, as shown in Table~\ref{tab:ablation}.
Please see supplementary material for the visualization of scaling maps and the shape of the predicted Gaussian points.

%% file: tables/camera_sparsity.tex
\newcommand{\tabledegree}{
\begin{table}
\caption{\textbf{Sensibility to camera sparsity.} We use the model trained under 8-camera setup to perform inference on a 6-camera setup.}
\small
\setlength\tabcolsep{2.0pt} 
\centering
\begin{tabular}{l|ccc|ccc}
\toprule
\multicolumn{1}{c}{\multirow{1}[6]{*}{Model}} & \multicolumn{3}{c}{8-camera setup} & \multicolumn{3}{c}{6-camera setup}\\

\cmidrule{2-7}   \multicolumn{1}{c}{} & PSNR$\uparrow$ & SSIM$\uparrow$ & \multicolumn{1}{c}{}{LPIPS$\downarrow$} & PSNR$\uparrow$ & SSIM$\uparrow$ & LPIPS$\downarrow$ \\ \midrule

FloRen~\cite{shao2022floren}     & 23.26 & 0.812 & 0.184 & 18.72 & 0.770 & 0.267\\
IBRNet~\cite{wang2021ibrnet} & 23.38 & 0.836 & 0.212 & 21.08 & 0.790 & 0.263\\
ENeRF~\cite{lin2022enerf} & 24.10 & 0.869 & 0.126 & 21.78 & 0.831 & 0.181\\
Ours & \textbf{25.57} & \textbf{0.898} & \textbf{0.112} & \textbf{23.03} & \textbf{0.884} & \textbf{0.168}\\

\bottomrule
\end{tabular}
\vspace{-2mm}
\label{tab:camera}
\end{table}
}

%% file: tables/ablation.tex
\newcommand{\tableablation}{
\begin{table}
\small
\caption{\textbf{Quantitative ablation study on synthetic data.} We report PSNR, SSIM and LPIPS metrics for evaluating the rendering quality, while the end-point-error (EPE) and the ratio of pixel error in 1 pix level for measuring depth accuracy.}
\vspace{-1mm}
\setlength\tabcolsep{2.5pt} 
\centering
\begin{tabular}{l|ccc|cc}
\toprule
\multicolumn{1}{c}{\multirow{1}[5]{*}{Model}} & \multicolumn{3}{c}{Rendering} & \multicolumn{2}{c}{Depth}\\

\cmidrule{2-6}   \multicolumn{1}{c}{} & PSNR$\uparrow$ & SSIM$\uparrow$ & \multicolumn{1}{c}{}{LPIPS$\downarrow$} & EPE $\downarrow$ & 1 pix $\uparrow$ \\ \midrule

Full model     & \textbf{25.05} & \textbf{0.886} & 0.121 & \textbf{1.494} & \textbf{65.94}\\
\textit{w/o} Joint Train. & 23.97 & 0.862 & \textbf{0.115} & 1.587 & 63.71\\
\textit{w/o} Depth Enc. & 23.84 & 0.858 & 0.204 & 1.496 & 65.87\\

\bottomrule
\end{tabular}
\vspace{-3mm}
\label{tab:ablation}
\end{table}
}

%% file: sections/6_conclusion.tex
\section{Discussion}
\label{sec:conclusion}
\noindent\textbf{Conclusion.}
By directly regressing pixel-wise Gaussian parameter maps defined on source view image planes, our GPS-Gaussian takes a significant step towards a real-time photo-realistic human novel view synthesis system under sparse-view camera settings. The proposed pipeline is fully differentiable and carefully designed. We demonstrate that our method notably improves both quantitative and qualitative results compared with baseline methods and achieves a much faster rendering speed on a single RTX 3090 GPU.

\noindent\textbf{Limitations.}
Although the proposed GPS-Gaussian synthesizes high-quality images, some elements still impact the effectiveness of our method.
For example, accurate foreground matting is necessary as a preprocessing step since we mainly focus on synthesizing the novel views of human performers.
Therefore, it is not straightforward to generalize our method to more general tasks.
Besides, the ground truth depths are required for supervision, increasing the difficulty of training data acquisition.
We believe that collecting massive high-quality synthetic data covering variant scenarios is conducive to alleviating these problems.

\noindent\textbf{Acknowledgement.}
This paper is supported by National Key R\&D Program of China (2022YFF0902200), the NSFC project (Nos. 62272134, 62236003, 62072141, 62125107 and 62301298), Shenzhen College Stability Support Plan (Grant No. GXWD20220817144428005) and the Major Key Project of PCL (PCL2023A10-2).

%% file: sections/X_suppl.tex
\clearpage
\setcounter{page}{1}
\maketitlesupplementary
\input{tables/runtime}

We present the visualization of opacity maps (Sec.~\ref{sec:vis_opacity}) and scaling maps (Sec.~\ref{sec:vis_scaling}), performance under randomly placed camera setup (Sec.~\ref{sec:random_setup}), run-time comparison (Sec.~\ref{sec:run-time}), network architecture (Sec.~\ref{sec:network}) and live demo setting (Sec.~\ref{sec:live}).

\section{Visualization of Opacity Maps}
\label{sec:vis_opacity}
\begin{figure}[h]
  \vspace{-1mm}
  \centering
  \includegraphics[width=0.475\textwidth]{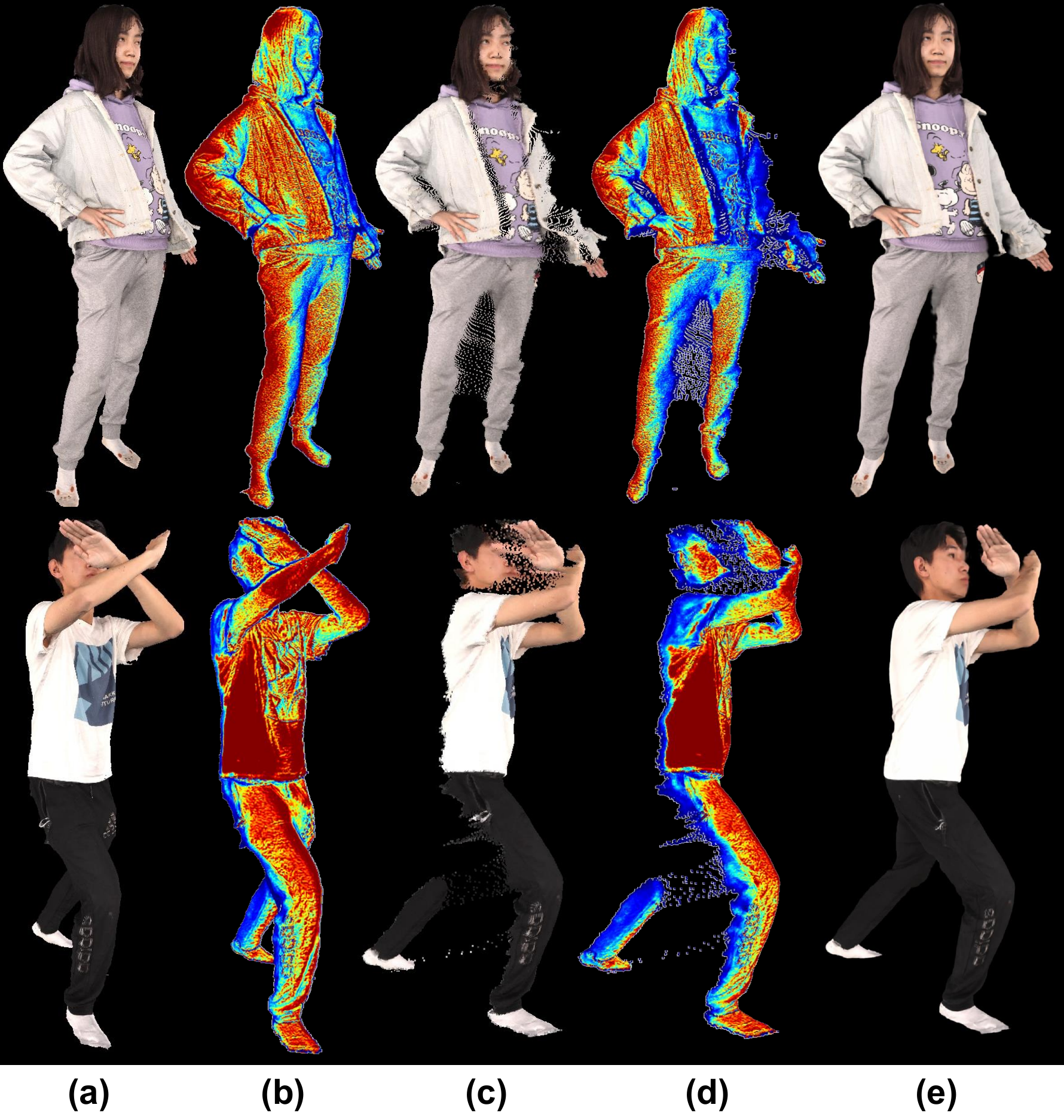}
  \vspace{-5mm}
  \caption{\textbf{Visualization of opacity maps.} (a) One of the source view images. (b) The predicted opacity map related to (a). (c)/(d) The directly projected color/opacity map at novel viewpoint. (e) Novel view rendering results. A cold color in (b) and (d) represents an opacity value near 0, while a hot color near 1. The low opacity values predicted for the outliers make them invisible.}
  \label{fig:vis_opacity}
\end{figure}

As mentioned in Sec.~\ref{sec:ablation}, the joint regression with Gaussian parameters eliminates the outliers by predicting an extremely low opacity for the Gaussian points centered at these positions.
The visualization of opacity maps is shown in Fig.~\ref{fig:vis_opacity}.
Since the depth prediction works on low resolution and upsampled to full image resolution, the drastically changed depth in the margin areas causes ambiguous predictions (\emph{e.g.} the front and rear placed legs of the girl and the crossed arms of the boy in Fig.~\ref{fig:vis_opacity}).
These ambiguities lead to rendering noise on novel views when using a point cloud rendering technique.
Thanks to the learned opacity map, the low opacity values make the outliers invisible in novel view rendering results, as shown in Fig.~\ref{fig:vis_opacity}~(e).
 
\section{Visualization of Scaling Maps}
\label{sec:vis_scaling}
\begin{figure}[t]
  \centering
  \includegraphics[width=\linewidth]{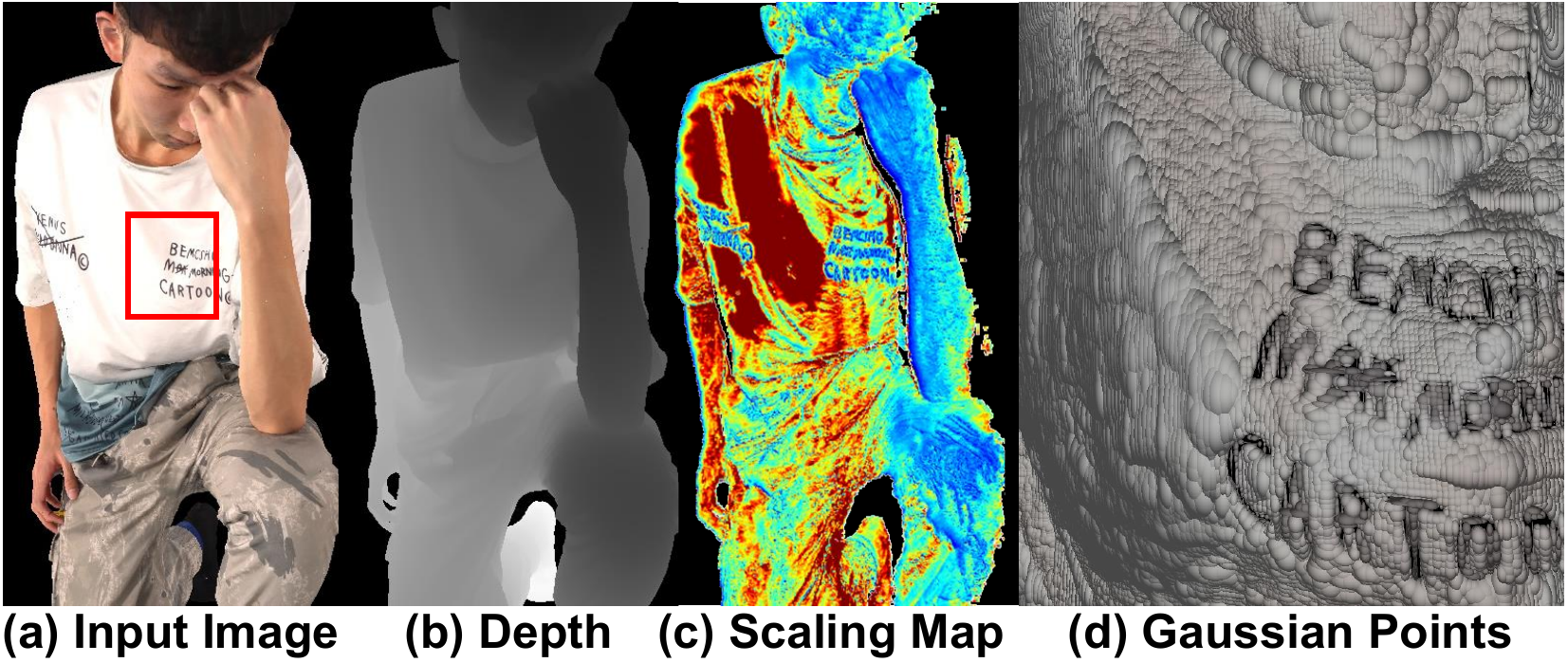}
  \vspace{-6mm}
  \caption{\textbf{Visualization of scaling map and the shape of Gaussian points.} (a) One of the source view images. (b) The depth of (a). (c) The scaling map shown in heat map, where a hotter color represents a larger value. (d) The zoom-in Gaussian points of the boxed area in (a). The depth and scaling map are normalized.}
  \label{fig:vis_scaling}
  \vspace{-5mm}
\end{figure}

The visualization of the scaling map (mean of three axes) in Fig.~\ref{fig:vis_scaling}~(c) indicates that the Gaussian points with lower depth roughly have smaller scales than the distant ones.
However, the scaling property is also impacted by comprehensive factors.
For example, as shown in Fig.~\ref{fig:vis_scaling}~(c) and (d), fine-grained textures or high-frequency geometries lead to small-scaled Gaussians.

\vspace{-1mm}
\section{Randomly Placed Camera Setup}
\label{sec:random_setup}
\begin{figure}[h]
  \centering
  \includegraphics[width=\linewidth]{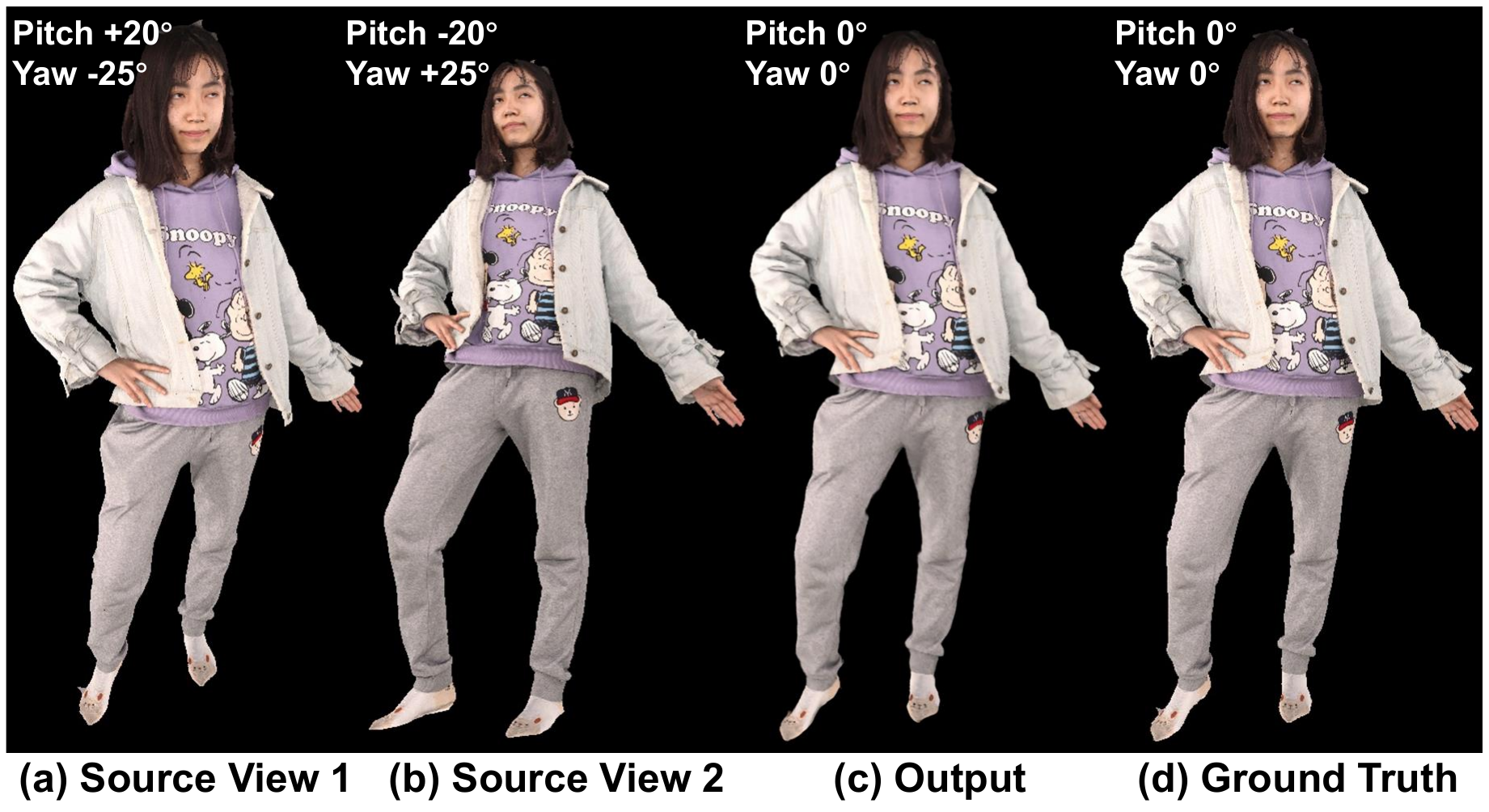}
  \vspace{-6mm}
  \caption{\textbf{Result on randomly placed camera setup.} (a) and (b) are the source view images with an extreme pitch and yaw. (c) is the novel view rendering result. (d) is the novel view ground truth.}
  \label{fig:pitching}
\end{figure}

We test our method with a randomly placed camera setup in Fig.~\ref{fig:pitching}.
The model trained under a uniformly placed 8-camera setup in Sec.~\ref{sec:experiments} shows a strong generalization capability to random camera setup with a pitch in range of [$-20^\circ, +20^\circ$] and yaw in range of [$-25^\circ, +25^\circ$].
However, rendering additional synthetic data covering more general camera setups to re-train the model is a better choice for achieving improved performance in such cases.

\vspace{-2mm}

\section{Run-time Comparison}
\label{sec:run-time}

We compare the run-time of the proposed GPS-Gaussian with baseline methods.
As illustrated in Table~\ref{tab:runtime}, the overall run-time can be generally divided into two parts which correlate to the source views and the desired novel view respectively.
The source view correlated computation in FloRen~\cite{shao2022floren} refers to coarse geometry initialization while the key components, the depth and flow refinement networks, operate on novel viewpoints.
IBRNet~\cite{wang2021ibrnet} uses transformers to aggregate multi-view cues at each sampling point aggregated to the novel view image plane, which is time-consuming.
ENeRF~\cite{lin2022enerf} constructs two cascade cost volumes on the targeted novel viewpoint, then predicts the novel view depth followed by a depth-guided sampling for volume rendering.
Once the target viewpoint changes, these methods need to recompute the novel view correlated modules.
However, the computation on source views dominates the run-time of GPS-Gaussian, which includes binocular depth estimation and Gaussian parameter map regression.
Given a target viewpoint, it takes only 0.8 ms to render the 3D Gaussians to the desired novel view.
This allows us to render multiple novel views simultaneously, which caters to a wider range of applications such as holographic displays.
Suppose that $n=10$ novel views are required concurrently, it takes our method $T = T_{src} + n \times T_{novel} = 35ms$ to synthesize, while $124ms$ for FloRen and $1261ms$ for ENeRF.

\tableruntime

\section{Network Architecture}
\label{sec:network}
\begin{figure}[t]
  \centering
  \includegraphics[width=\linewidth]{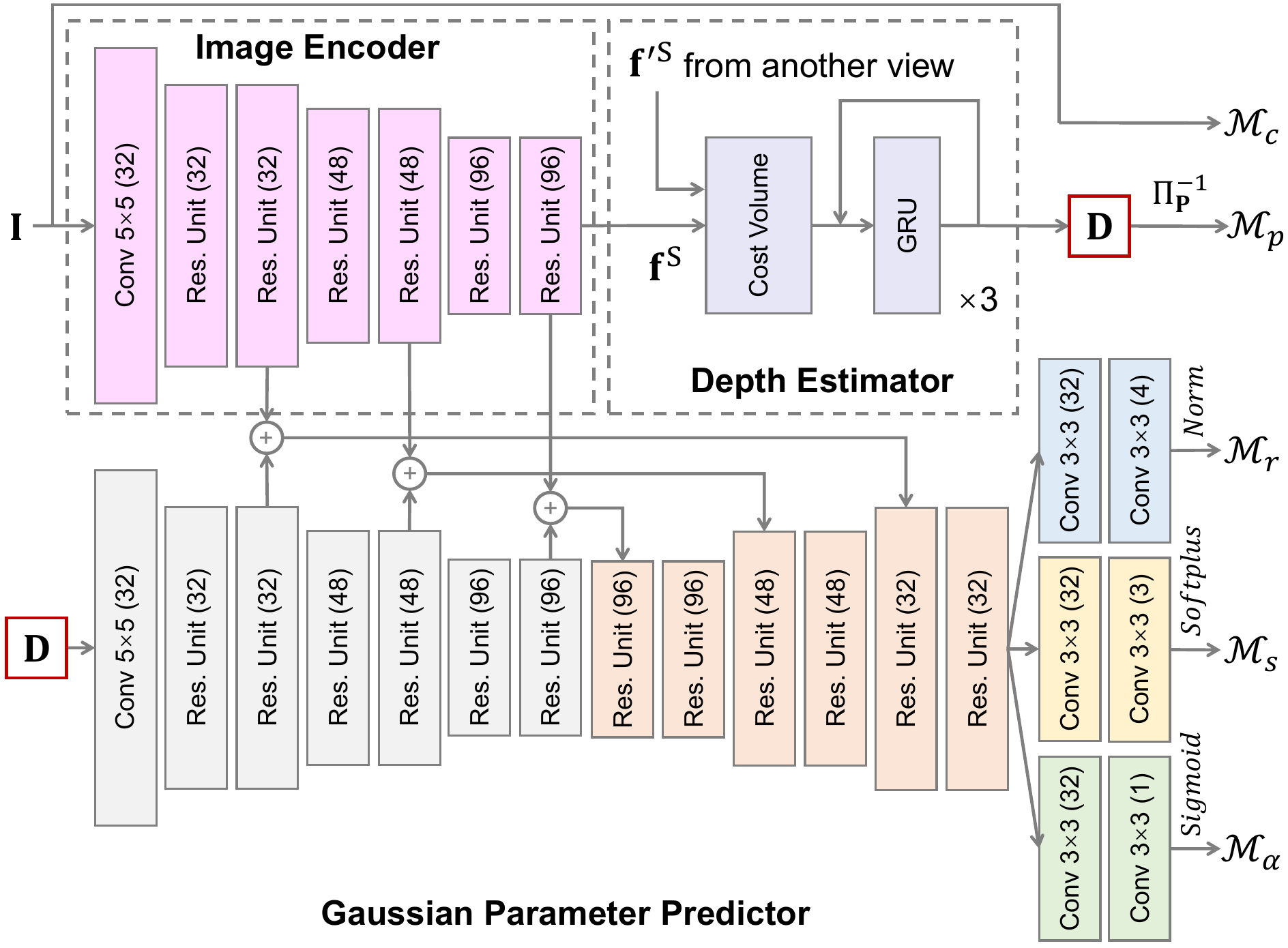}
  \vspace{-5mm}
  \caption{\textbf{Network architecture.} The proposed framework takes a source image as input to regress Gaussian parameter maps.}
  \label{fig:network}
  \vspace{-4mm}
\end{figure}

As shown in Fig.~\ref{fig:network}, the network architecture of the proposed GPS-Gaussian is composed of (1) image encoder, (2) depth estimator, and (3) Gaussian parameter predictor.

\noindent\textbf{Image Encoder.}
The image encoder $\mathcal{E}_{img}$ is applied to both source images and maps each of them to a set of dense feature map $\{\mathbf{f}_l^s\}_{s=1}^S$ as in Eq.~\ref{formula:image encoder}.
$\mathcal{E}_{img}$ has a similar architecture to the feature encoder in RAFT-Stereo~\cite{lipson2021raft-stereo}.
We change the kernel size of the first convolution layer from $7\times7$ to $5\times5$ and replace all batch normalization with group normalization.
Residual blocks and downsampling layers produce image features in 3 levels at $1/2$, $1/4$ and $1/8$ the input image resolution, with $32$, $48$ and $96$ channels, respectively. 
The extracted features are further used to construct the correlation volume and regress the Gaussian parameters.

\noindent\textbf{Depth Estimator.}
As mentioned in Sec.~\ref{sec:selection&depth}, The classic binocular stereo methods only estimate the depth for `reference view'.
while the `target view' feature is only used to construct the cost volume but is not involved in further depth estimation or iterative refinement.
By making the image encoder independent from the depth estimator, and re-indexing the correlation volume $\mathbf{C}$ for both lookup procedures, we realize a compact and parallelized implementation that results in a decent efficiency increase exceeding 30\%.
For the refinement module, we set $T=3$ considering the trade-off between the performance and the cost.

\vspace{1mm}
\noindent\textbf{Gaussian Parameter Predictor.}
This module is composed of a depth encoder $\mathcal{E}_{depth}$ and a U-Net like Gaussian parameter decoder $\mathcal{D}_{parm}$.
$\mathcal{E}_{depth}$ takes the predicted depth as input and has an identical architecture to the image encoder.
Image features concatenated with depth features are aggregated to the Gaussian parameter decoder via skip connections.
The decoded pixel-wise Gaussian feature $\mathbf{\Gamma}$ passes through three specific prediction heads to get rotation map $\mathcal{M}_r$, scaling map $\mathcal{M}_s$ and opacity map $\mathcal{M}_\alpha$, respectively.
Meanwhile, the position map $\mathcal{M}_p$ is determined by the predicted depth map $\textbf{D}$ and the color map $\mathcal{M}_c$ directly borrows from the RGB value of the input image.

\section{Live Demo Setting}
\label{sec:live}
We present live demos on our project page, in which we capture source view RGB streams and synthesize novel views in one system.
Due to the memory limit of RTX 3090 GPU, we connect the front 6 cameras (facing human subjects) to the computer, which are uniformly positioned in a circle of a 2-meter radius. 
GPS-Gaussian enables real-time high-quality rendering, even for challenging hairstyles and human-object or multi-human interactions.
For a more in-depth exploration of our results, please visit our homepage:  {\small\url{shunyuanzheng.github.io/GPS-Gaussian}}.

%% file: tables/runtime.tex
\newcommand{\tableruntime}{
\begin{table}[!htbp]
\caption{\textbf{Run-time comparison.} We report the run-time correlated to the source views and each novel view on an RTX 3090 GPU. All methods take two $1024\times1024$ source images as input. Our method can render multiple novel views concurrently in real-time.}
\vspace{-2mm}
\small
\setlength\tabcolsep{1pt} 
\centering
\begin{tabular}{l|cc}
\toprule 
Methods & Source view  & Novel view (per view) \\ 
\midrule
FloRen~\cite{shao2022floren}  & 14 ms & 11 ms   \\ 
IBRNet~\cite{wang2021ibrnet}  & 5 ms  & 4000 ms \\
ENeRF~\cite{lin2022enerf}     & 11 ms & 125 ms  \\
Ours                          & 27 ms & 0.8 ms  \\
\bottomrule

\end{tabular}
\vspace{-6mm}
\label{tab:runtime}
\end{table}
}